\documentclass{IEEEtran}

\title{Data Augmentation and Hyperparameter Tuning for Low-Resource MFA}

\author{Alessio Tosolini and Claire Bowern \\ Department of Linguistics\\ Yale University, New Haven, CT}

\newcommand{\blue}[1]{\textcolor{blue}{#1}}
\usepackage{comment}
\usepackage{todonotes}
\usepackage{url}
\usepackage[square,numbers]{natbib}
\begin{document}

\maketitle

\begin{abstract}
A continued issue for those working with computational tools and endangered and under-resourced languages is the lower accuracy of results for languages with smaller amounts of data. We attempt to ameliorate this issue by using data augmentation methods to increase corpus size, comparing augmentation to hyperparameter tuning for multilingual forced alignment. Unlike text augmentation methods, audio augmentation does not lead to substantially increased performance. Hyperparameter tuning, on the other hand, results in substantial improvement without (for this amount of data) infeasible additional training time. For languages with small to medium amounts of training data, this is a workable alternative to adapting models from high-resource languages.
    
    

\end{abstract}

\section{Introduction}
A continued issue for those working with computational tools and endangered and under-resourced languages is worse model performance, resulting in lower accuracy of results, for languages with smaller amounts of data. Materials created in low-resource circumstances are disproportionately created outside controlled lab settings. That is, as well as there being less data to work with, the data that is there may be more difficult to work with. Various other work has examined how to increase performance in such cases, including using or adapting models created on high-resource languages \cite{wang2023evaulating,dicanio_using_2013,sanLeveragingPreTrainedRepresentations2021,johnsonForcedAlignmentUnderstudied2018}. 


This experiment explores two previously unexplored methods for better low-resource forced alignment using the Montreal Forced Aligner (MFA; \cite{mcauliffeMontrealForcedAligner2017}): augmenting the training data and finding optimal hyperparameters through hyperparameter tuning. Data augmentation is the process of increasing the data available for model training by generating modified copies of the original data. Data augmentation is widely used in text-based NLP \cite{LI202271} and computer vision \cite{Shorten2019ASO}, and has been shown to improve model performance in contexts where annotated training data is scarce \cite{chen2021empiricalsurveydataaugmentation}. Training a model on augmented datasets has provided substantial improvements in model accuracy for a variety of tasks, including machine translation \cite{oh2023dataaugmentationneuralmachine} and hate detection \cite{khullar2023hatespeechdetectionlimited}. The literature for audio data augmentation is more scarce, but data augmentation strategies have led to improved models for automatic speech recognition \cite{9414483} and speech emotion recognition \cite{ahmed2022ensemble1dcnnlstmgrumodeldata}, among others. Within text-based NLP, data augmentation might include text manipulation that retains close characteristics of natural language (e.g. replacing a word for its synonym \cite{kolomiyets-etal-2011-model,Sahin2021To} or shuffling phrases in a language that allows for multiple word orders \cite{takahagi-shinnou-2023-data}); other methods do not result in grammatical text (like randomly inserting/deleting a character or swapping words for a random other word) but nonetheless also lead to model improvements \cite{Sahin2021To}.

Hyperparameter tuning is the process of searching a space of model parameter values to maximize accuracy or minimize loss. Hyperparameter tuning is common when training neural networks but can be applied to non-neural contexts too. In this paper, we explore the hyperparameters that relate to each of the four stages of training: monophone, triphone, linear discriminant analysis (LDA), and speaker-adapted training (SAT). Learning monophone alignments consists of learning how to model individual phones. Triphone alignments models phones in the context of the preceding and following phones. By default, each phone is processed uniquely during the learning of triphone alignments, but a model may optionally be trained with phone groupings that collapse distinctions between phones preceding and following the target phone during the alignments phase. Learning LDA alignments involves finding an optimal reduction of acoustic features as to maximise the distinctions between phone classes. SAT normalizes the differences between speakers as to diminish the effects of the individual speaker on alignments.

\section{Methodology}
The methodology for this paper largely follows the procedures described in Tosolini \& Bowern (2025) \cite{tosolinibowern2025} and uses much of the same data. We use a subset of  collections compiled from public language archives for analyses reported in \cite{babinskiArchivalPhoneticsProsodic2022}; see Table~\ref{tab:corpus}. The total amount of training data comprises 10 hours of data from 6 Australian Indigenous languages. Files were downloaded from archives and segmented at the word and segment level using \texttt{p2fa}, the local version of the FAVE forced aligner \citep{rosenfelder2011fave} (requiring a g2p based on the English arpabet). Praat \citep{praat} was then used to manually correct misplaced boundaries. This is the dataset against which we compare the accuracy of automatically aligned files.

\begin{table*}[t!]
    \centering
    \begin{tabular}{lllp{4.5cm}c}
    \hline
      Language & Language Family & Reference   & Collector & Minutes \\ \hline
      Bardi & Nyulnyulan & A: Bowern\underline{ }C05 & Claire Bowern  & 108 \\
      Gija & Jarrakan & E: 0098MDP0190 & Frances Kofod & 157 \\
      Kunbarlang & Gunwinyguan & E: 0384SG0324 & Isabel O'Keefe; Ruth Singer & 16\\
      Ngaanyatjarra & Pama-Nyungan & P: WDVA1 & Inge Kral & 53 \\
      Yan-nhangu & Pama-Nyungan & E: dk0046 & Claire Bowern & 290\\
      Yidiny & Pama-Nyungan & A: A2616 & R.M.W.\ Dixon & 50\\
  \hline
    \end{tabular}
    \caption{Corpus information. A: AIATSIS; E: Elar; P: Paradisec}
    \label{tab:corpus}
\end{table*}

The transcriptions for these materials were done by the original researchers and include various artefacts typical of corpora made for language documentation and use by humans, rather than computational processing. We therefore normalized a number of features and removed some data to allow processing for augmentation and forced alignment. This included removing partially transcribed words, removing notes from transcription tiers (so that only material in the audio record is in a transcript), and standardizing some aspects of transcription (such as removing hyphens). We did not otherwise alter transcriptions. We also removed all intervals shorter than 100ms. We did not further review transcripts for accuracy.
Model training used two datasets. One is the \textsc{Yidiny-Train} corpus: 38 minutes of Yidiny narratives. The other is the \textsc{Big5} dataset (646 minutes), comprised of the entirety of the Bardi, Gija, Ngaanyatjarra, and YanNhangu materials as well as the Yidiny-Train corpus. Audio augmentation techniques were used to increase the amount of training data that the models were trained on. Preliminary experiments involved training and evaluating models on training datasets consisting of the unaugmented data and one of twelve modifications listed in Table~\ref{table:Augmentations}, alongside an unmodified dataset. All augmented data were created with the pydub \cite{robert2018pydub}, librosa \cite{mcfee2015librosa}, or noisereduce \cite{sainburg2019noisereduce} packages. Of the twelve modifications, four of them (a bassboost by a factor of 2, a low-pass filter of 4000Hz, converting the wav file to mp3 and back with a sampling rate of 128k, and slowing the file down by 20\%) resulted in slight improvements over a model trained on an unaugmented dataset and the control of twice the data. Thus, we created the AugBig5 and AugYidiny training datasets by taking the Big5 and Yidiny dataset respectively and generating a dataset with all four data augmentation strategies and the original data. 

\begin{table}[]
\begin{tabular}{llp{4.25cm}}
\hline
Augmentation &   Incl.\ & Explanation \\
\hline
default  &  Yes & No change \\
bassboost &  Yes  & Bass boost by factor of 2\\
downsample & No & downsample to 8,000 Hz\\
f0change &  No &  increase f0 by 20\% \\
f0change &  No &  reduce f0 by 20\%\\
intensitychange & No & halve intensity\\
lowpassfilter & Yes & remove signal above 4000Hz\\
noisereduce  & No & Use noisereduce filter\\
wavtomp3towav &Yes & convert to mp3 at 128k, then back to wav (16kHz)\\
wavtomp3towav & No & convert to mp3 at 64k, then back to wav (16kHz)\\
shift & No & shift all boundaries by 5ms \\
speedchange  & Yes & reduce duration of all intervals by 20\% \\
speedchange  & No & increase duration of all intervals by 20\% \\ \hline
\end{tabular}
    \caption{Types of audio manpulation for default augmentation}    \label{table:Augmentations}
\end{table}

\subsection{Acoustic Models}
All acoustic models in this experiment were trained from scratch using MFA's train command. Two experiments were conducted where the training datasets and the training hyperparameters were altered. The first experiment investigated the optimal hyperparameters for the type of training data, the categories defined in the triphone groupings, and number of training configurations. Figure~\ref{table:HyperparameterTable1} shows the modifications for each of the three parameters. This gave us a total of 208 models for the first experiment. 

The second experiment isolated the effects of modifying the configuration files by training seven models. The first model used the number of iterations included in the Montreal Forced Aligner configuration test\footnote{\url{https://github.com/MontrealCorpusTools/Montreal-Forced-Aligner/blob/main/tests/data/configs/basic_train_config.yaml}}: 5 monophone iterations, 3 triphone iterations, 2 LDA iterations, and 2 SAT iteartions. Six additional models were trained, each with twice the number of iterations in all four training categories as the previous one.

\begin{table}[]
\begin{tabular}{lll}
\hline
Controlled Parameter & Categories & Quantity \\ \hline
Training Dataset & \begin{tabular}[c]{@{}l@{}}AugBig5\\ Big5\\ AugYidiny\\ Yidiny\end{tabular} & 4 \\ \hline
Triphone Groupings & \begin{tabular}[c]{@{}l@{}}4-classes    \\ 9-classes    \\ 22-classes \\ default-classes\end{tabular} & 4 \\ \hline
Iterations & \begin{tabular}[c]{@{}l@{}}\{$\frac12$, 2, 4\}-monophone\\ \{$\frac12$, 2, 4\}-triphone\\ \{$\frac12$, 2, 4\}-LSA\\ \{$\frac12$, 2, 4\}-SAT \\ default\end{tabular} & 13 \\ \hline
\end{tabular}
    \caption{Hyperparameters investigated in Experiment 1}
    \label{table:HyperparameterTable1}
\end{table}

MFA requires a dictionary. We created a single dictionary by converting all word types in each languages to IPA and then combining them in a single file. Although not all languages have identical inventories, they all use 1:1 phonemic orthographies. Dictionary compilation was thus straightforward.
\subsection{Evaluation}
Unlike previous work by Tosolini \& Bowern (2025) \cite{tosolinibowern2025}, we evaluate models on only one testing setting: Yidiny-seen, corresponding to all the training data from the Yidiny language. Thus, for all models trained on exclusively Yidiny data, testing is performed on the training data. For the Big5 dataset, testing is performed on a subset of its training data. The authors of this paper decided to not include evaluations on held out testing data from Yidiny or other unseen languages because in most situations involving forced alignment for very low-resource languages, the purpose of a model is not to generalize to more data, but instead develop the best model possible to facilitate analyses with the data the researchers already have. For (near-)extinct languages, the data the researchers have will often be all the data they will ever have. This means generalizability is less of a priority for many linguistics compared to accuracy on seen data, so we thus only present testing data on training datasets. 

In experiment 1, the accuracies presented for the models with default parameters and default classes is not equivalent to a model that was not passed in any parameters and no triphone groupings. In order to ensure that the total number of iterations is 40 for a model trained on 5 data augmentation strategies, the ``default parameter"s for a model were divided by 5. This means that ``default parameter" is not equivalent to passing no configuration file. Similarly, to ensure consistency across triphone grouping files, models marked with ``default classes" were trained by passing in a triphone grouping file where each phone was in its own class. This is not equivalent to not passing in any triphone grouping file. Although both of these defaults change the accuracy, they were necessary to allow for comparison across different model types.

\section{Results}
\subsection{Experiment 1}

\begin{figure}
    \centering
    \includegraphics[width=\linewidth]{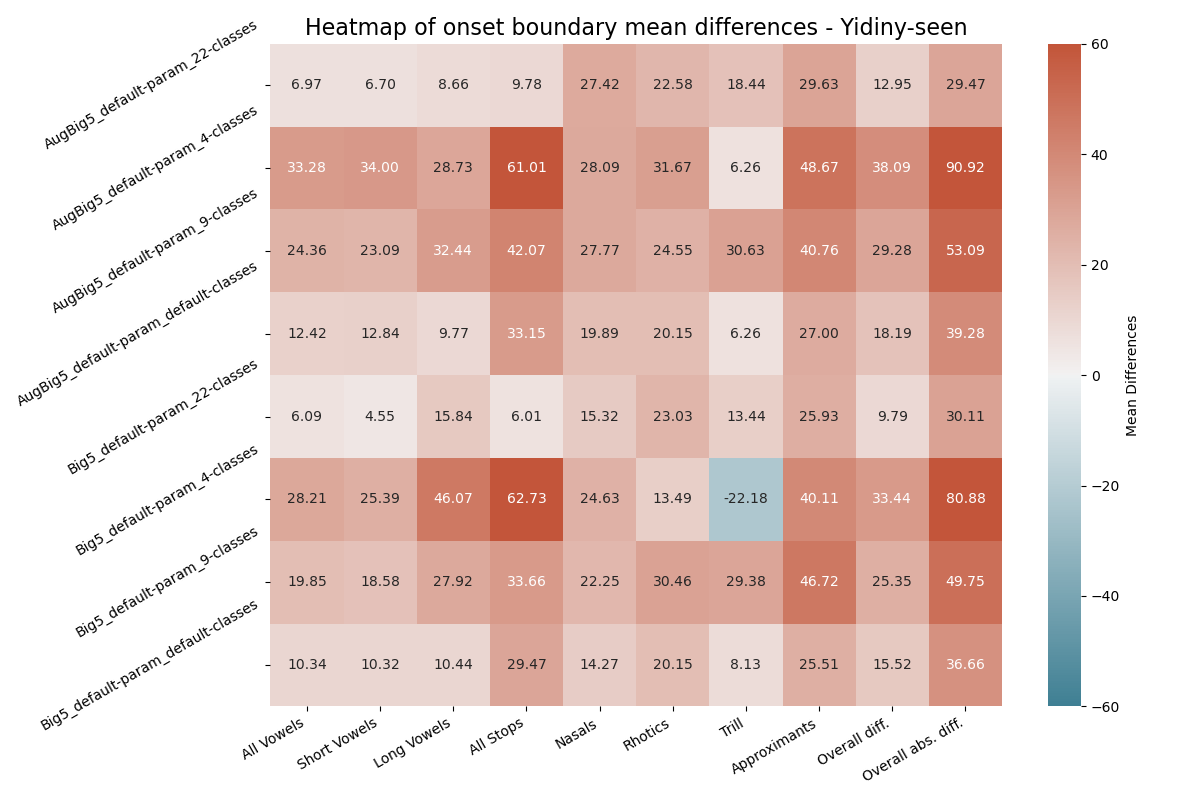}
    \caption{Mean differences for AugBig5 and Big5 datasets for different triphone groupings.}\vspace*{-0.5cm}
    \label{fig:exp1-triphone-groupings-Big5}
\end{figure}

We evaluate the results of experiment 1 across all four training datasets (AugBig5, Big5, AugYidiny, and Yidiny). We first investigate the effect of the number of triphone groupings on the accuracy of the models, keeping all other training parameters the same. Figure~\ref{fig:exp1-triphone-groupings-Big5} shows the effects of the number the triphone groupings on the mean accuracy across natural classes for the AugBig5 and Big5 datasets. For both the AugBig5 and Big5 datasets, the lowest mean absolute difference occurs for models trained with 22 phone classes. Additionally, across the board models trained on either dataset show that the average boundary placed by MFA occurs consistently after the average boundary placed by a human annotator, as seen with the red color of all the cells on the heatmap.

\begin{figure}
    \centering
    \includegraphics[width=1\linewidth]{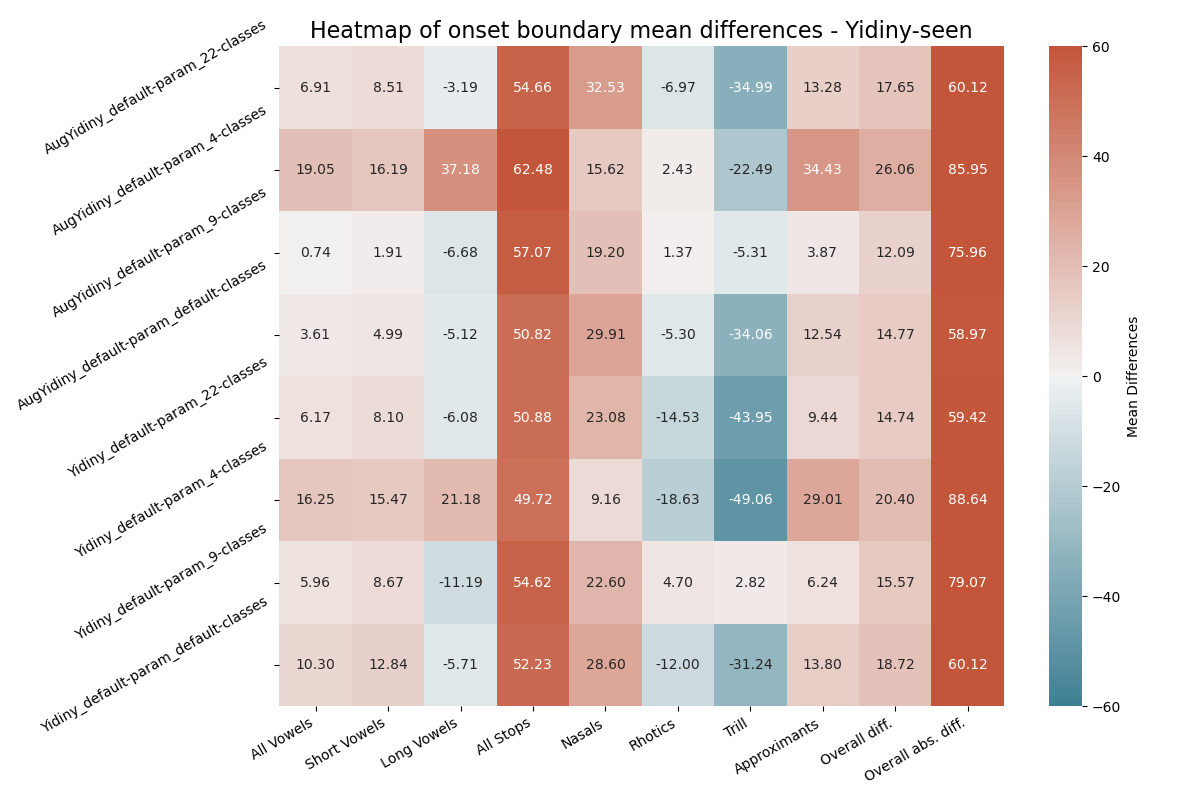}
    \caption{Mean differences for AugYidiny and Yidiny datasets for different triphone groupings.}
    \label{fig:exp1-triphone-groupings-Yidiny}
\end{figure}

Figure~\ref{fig:exp1-triphone-groupings-Yidiny} similarly shows the effects of the triphone groupings on the AugYidiny and Yidiny datasets. Note that unlike for the AugBig5 and Big5 datasets, creating phone groupings does not necessarily result in a large increase in mean accuracy, with the AugYidiny dataset showing the highest overall absolute accuracy for the model trained with each phone in its own class. However, the difference seems to be minimal. Interestingly, we also no longer see the trend of the MFA's boundaries being consistently placed after the human-annotated boundaries across the board, with the error seeming to rely more heavily on the natural class of the natural class of the phone, with long vowels, rhotics, and trills having a negative difference between the MFA's predicted boundary and the human-annotated boundary, indicating that the MFA's boundary is placed before the human-annotated boundary.

\begin{figure}
    \centering
    \includegraphics[width=1\linewidth]{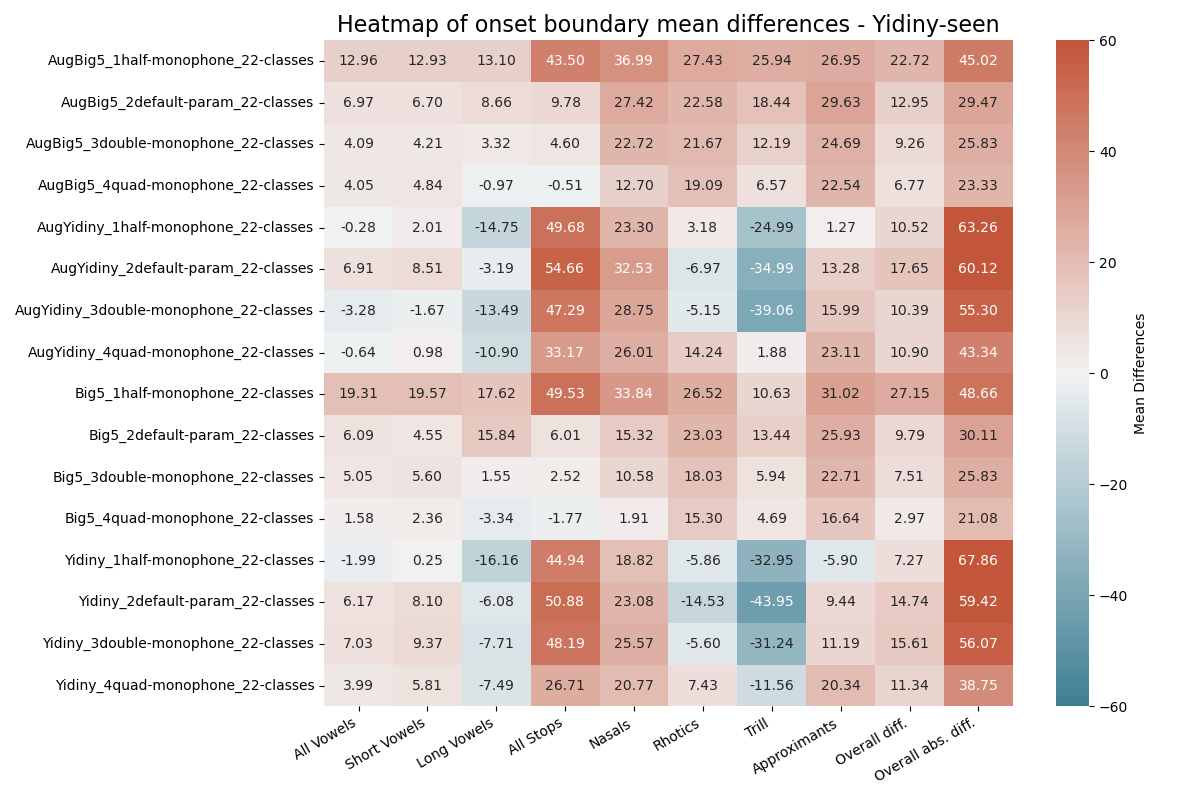}
    \caption{Mean differences for different amounts of monophone iterations.}
    \label{fig:exp1-monophone}
\end{figure}

Due to the near categorical benefits of using a 22-class triphone grouping across all four testing settings, future plots will only compare models that were trained with the 22-class triphone grouping. Figure~\ref{fig:exp1-monophone} compares models trained on every dataset for every number of monophone iterations. Across all models, there is a consistent trend favoring more monophone iterations. Higher monophone iterations allows for better boundaries for natural classes that are challenging for the models to find, such as the boundaries for stops and trills for AugYidiny- and Yidiny-based models. 

Unlike the modifications of number of monophone iterations, there was to be no major or consistent improvement from changing the number of triphone, LDA, or SAT iterations. None of the models from experiments targeting triphone, LDA, or SAT iterations gave mean overall absolute differences less than 25ms, suggesting that of the four configuration hyperparameters tested, the most significant for model accuracy is the number of monophone iterations.




Across all settings, augmenting the training data seemed to have a minimal effect. The models that achieved the lowest absolute difference between the human-annotated boundaries and the MFA's automatically placed boundaries were the AugBig5 and Big5 models with quad-monophone configuration hyperparameters and 22 phone classes, with scores of 23.33ms and 21.08ms respectively. Again, both of these models outperformed the best models trained from scratch in previous work with the same datasets \cite{tosolinibowern2025}, which trained on the Big5 dataset and achieved an accuracy of 23.92ms. Across all hyperparameters we explored, models trained on the AugBig5 and Big5 dataset much outperformed the models trained on the AugYidiny and Yidiny dataset, contradicting previous findings \cite{tosolinibowern2025}  that multilingual training datasets only slightly increase accuracy. We explore why this might be the case in the discussion.

\subsection{Experiment 2}

\begin{figure}
    \centering
    \includegraphics[width=\linewidth]{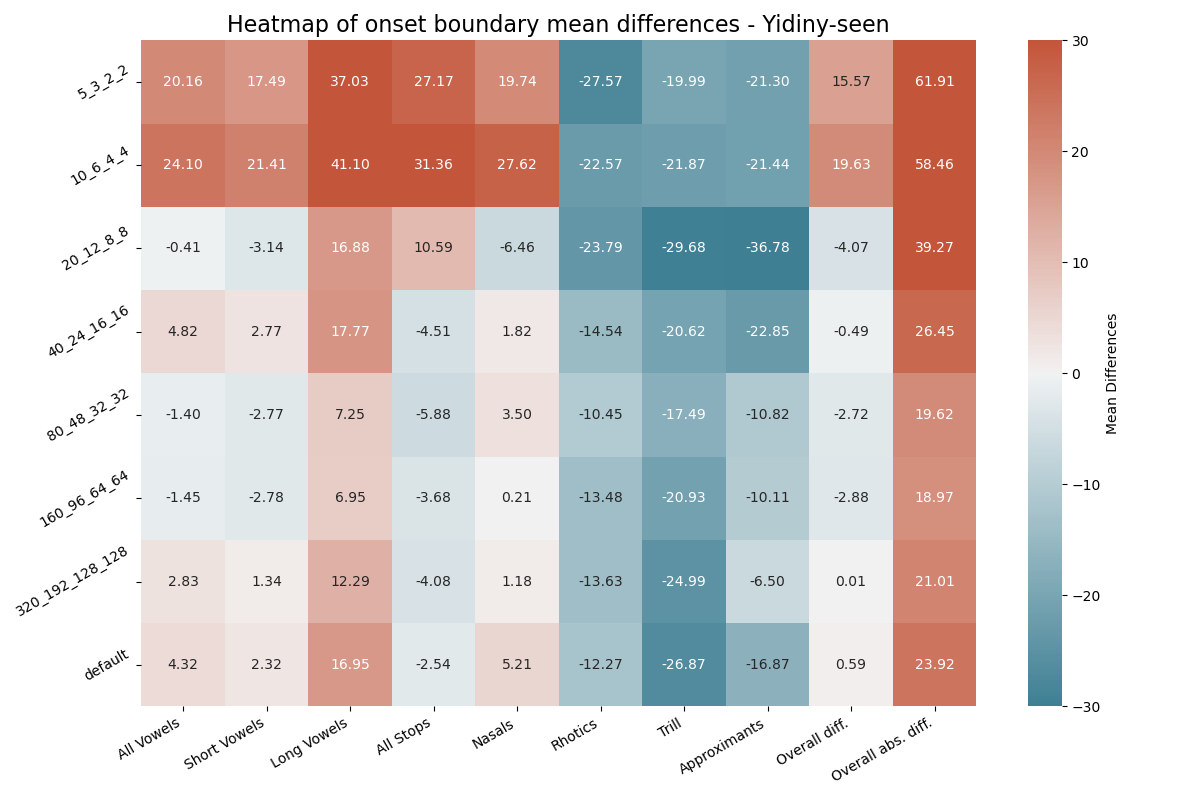}
    \caption{Mean differences (Big5 models) across  configurations}
    \label{fig:exp2-Big5}\vspace*{-0.25cm}
\end{figure}

\begin{figure}
    \centering
    \includegraphics[width=\linewidth]{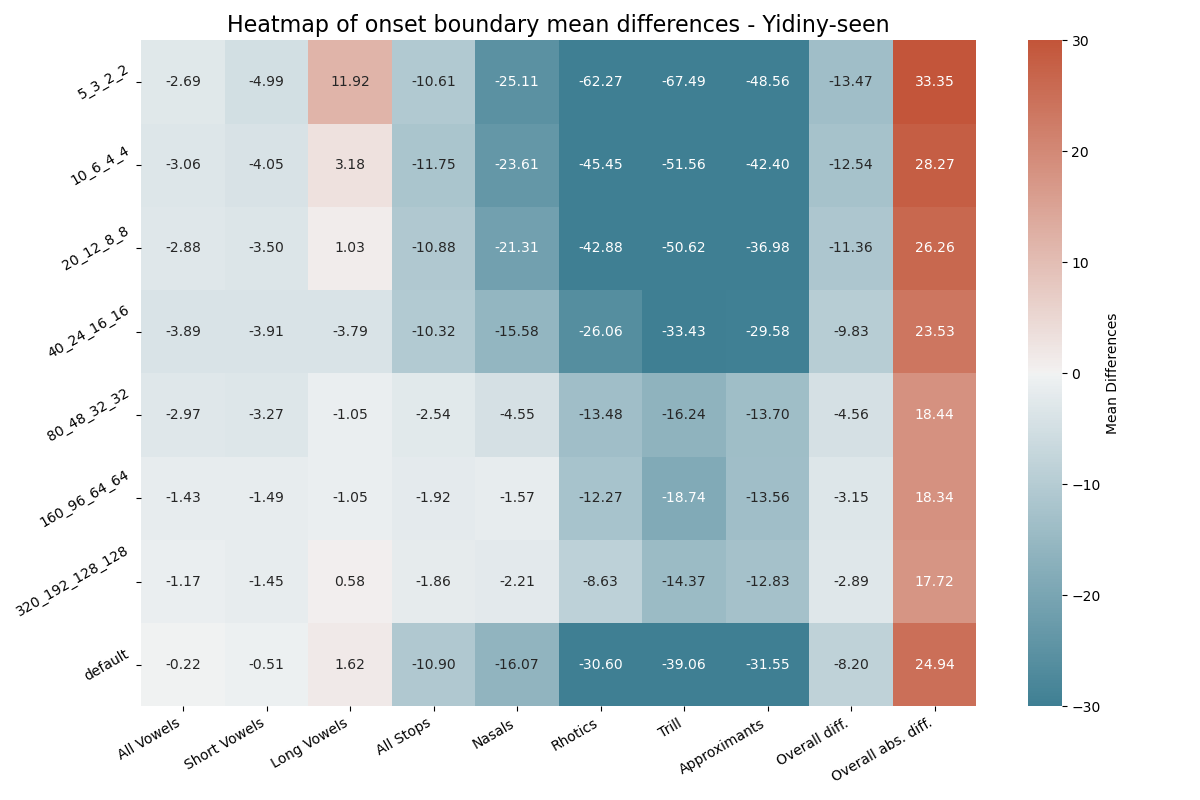}
    \caption{Mean differences (Yidiny) across configurations}
    \label{fig:exp2-Yidiny}\vspace*{-0.25cm}
\end{figure}

In the second experiment, we isolate the effects of the configurations by training models only on the Big5 and Yidiny datasets while varying the number of monophone, triphone, LSA, and SAT iterations such that the ratio of each iteration to each other remains constant.  Figures~\ref{fig:exp2-Big5} and~\ref{fig:exp2-Yidiny} show the effects of varying the number of iterations. Row labels represent the number of monophone, triphone, LSA, and SAT iterations, while the bottom row shows a model trained with no additional configurations specified.\footnote{The top column, 5\_3\_2\_2 is a copy of the basic training configuration found at https://github.com/MontrealCorpusTools/Montreal-Forced-Aligner/blob/main/tests/data/configs/basic\_train\_config.yaml. Subsequent columns change double the value of num\_iterations for each of the four training stages.}

Across both the models trained on the Big5 and Yidiny datasets, increasing the number of iterations results in better accuracies, as can be seen by the rightmost column generally decreasing with each row. Both the Big5 and Yidiny models consistently struggle with placing boundaries for rhotics, trills, and approximants;  these natural classes  benefit most from increased iterations. In this experiment, we see similar results to previous work on multilingual MFA alignment \cite{tosolinibowern2025}, in that multilingualism does not seem to play a role in increasing model performance. In fact, the best performing model across both experiments and datasets is the 320\_192\_128\_128 Yidiny model. 

\section{Discussion}

The best MFA acoustic models are generally achieved by taking a high resource model (e.g. from English), and then using it in a low-resource context, either with or without further adaptation \cite{tosolinibowern2025, chodroffComparingLanguagespecificCrosslanguage2024}. Additionally, training a monolingual model from scratch on a low-resource language can give similar accuracies as the best English based models if trained on around 80 minutes of audio \cite{chodroffComparingLanguagespecificCrosslanguage2024}. In both experiments presented in this paper, we find that models trained from scratch can achieve accuracies on seen datasets similar to the best fine-tuned models with only 30 minutes of training data, given custom training configurations, reducing the amount of audio data needed for the best models by 50 minutes or over 60\%.

The first experiment tested the effects of augmenting the training data, setting triphone groupings, and changing the number of iterations of each variable. Across all settings, accuracy was consistently higher for the AugBig5 and Big5 models compared to the AugYidiny and Yidiny models, suggesting that multilingual models may outperform monolingual models in certain settings. However, due to default configuration parameters not being equivalent to passing no configuration parameters, we are unable to truly compare the models trained on multilingual and monolingual datasets. We find no consistent increase in improvement due to training on augmented datasets. Instead, augmenting the datasets generally results in a slight decrease in performance compared to the the non-augmented datasets. 
Models which were trained with the optional triphone grouping parameter filled yielded highest accuracies with 22 phone classes. 
Lastly, increasing monophone iterations seemed to play the largest role in higher model accuracy. Increasing triphone, LSA, or SAT iterations did not show a clear increase in accuracy, with the number of monophone iterations having the largest role in model accuracy. 

The second experiment isolated the effects of increasing the number of iterations for all four states of training. This experiment demonstrated that simply by increasing the amount of iterations during training, model accuracy can get within 1ms of the best English-adapted models on a seen testing setting with only 30 minutes of training data. Note that the higher accuracy of the 40\_24\_16\_16 model compared to the default model (which would be depicted as a 35\_40\_40\_40 model due to MFA train's default parameters) hints at possible future directions for this work. Even a slight increase in monophone alignments may result in a better model, underscoring the importance of the number of monophone iterations for training an MFA model from scratch. 

These results highlight the efficacy of configuring hyperparameters and triphone groupings to increase model accuracy in low-resource settings, opening up the possibility of high performing models with as little as 30 minutes of training data. All experiments in this paper focus on models where overfitting is encouraged to best replicate the environments in which alignment is most often used by field researchers documenting endangered languages.


\section{Acknowledgements}

    The authors would like to thank the members of the Yale language contact and phonology groups for feedback on this project.


\bibliography{mybib,references}

\end{document}